\documentclass{article}

\PassOptionsToPackage{numbers, compress}{natbib}

\usepackage[preprint]{neurips_2023}

\usepackage[utf8]{inputenc} 
\usepackage[T1]{fontenc}    
\usepackage{url}            
\usepackage{booktabs}       
\usepackage{amsfonts}       
\usepackage{nicefrac}       
\usepackage{microtype}      
\usepackage{graphicx}

\usepackage{amssymb}
\usepackage{bm}
\usepackage{overpic}
\usepackage{multicol}
\usepackage{multirow}
\usepackage{siunitx}
\usepackage{enumitem}
\usepackage{colortbl}
\usepackage[dvipsnames,table,xcdraw]{xcolor}

\usepackage{pifont}

\newcommand{\figref}[1]{Fig.~\ref{#1}}
\newcommand{\tabref}[1]{Tab.~\ref{#1}}

\newcommand{\secref}[1]{Sec.~\ref{#1}}
\newcommand{\myPara}[1]{\vspace{6pt}\noindent\textbf{#1}}

\usepackage{makecell}

\usepackage[pagebackref,breaklinks,colorlinks,linkcolor=BrickRed,citecolor=RoyalBlue]{hyperref}

\def\MyMthd{MaskDiffusion}

\graphicspath{{./figures/}}
\usepackage{amsmath}
\usepackage{algorithm}
\usepackage{algorithmic}
\title{\MyMthd{}: Boosting Text-to-Image Consistency with Conditional Mask}

\author{Yupeng Zhou$^1$ \quad Daquan Zhou$^2$ \quad Zuo-Liang Zhu$^1$   \quad Yaxing Wang$^1$ \\ \textbf{ \quad Qibin Hou$^1$\thanks{Corresponding author.} \quad Jiashi Feng$^2$}\\ \\
$^1$VCIP, School of Computer Science, Nankai University\\
$^2$ByteDance, Singapore\\
{\tt\small \{ypzhousdu, andrewhoux\}@gmail.com} \quad
}

\begin{document}

\maketitle

\begin{abstract}
Recent advancements in diffusion models have showcased their impressive capacity to generate visually striking images.
Nevertheless, ensuring a close match between the generated image and the given prompt remains a persistent challenge.
In this work, we identify that a crucial factor leading to the text-image mismatch issue is the inadequate cross-modality relation learning between the prompt and the output image.
To better align the prompt and image content, we advance the cross-attention with
an adaptive mask, which is conditioned on the attention maps and the prompt embeddings, to dynamically adjust the contribution of each text token to the image features.
This mechanism explicitly diminishes the ambiguity in semantic information embedding from the text encoder, leading to a boost of text-to-image consistency in the synthesized images.
Our method, termed \MyMthd{}, is training-free and hot-pluggable for popular pre-trained diffusion models. 
When applied to the latent diffusion models, 
our \MyMthd{} can significantly improve the text-to-image consistency with negligible computation overhead compared to the original diffusion models. 

\end{abstract}

\section{Introduction}

Diffusion models~\cite{ho2021cascaded, NEURIPS2021_49ad23d1, yang2022diffusion, zhou2023magicvideo, xu2022versatile} have been the most prevailing methods amongst generative methods, revealing its superiority in a variety of down-streaming tasks~\cite{brack2022stable, chen2023fantasia3d, raj2023dreambooth3d, kumari2022customdiffusion, yang2022paint, kawar2023imagic, wang2022diffusiongan}. 
Recent state-of-the-art text-to-image (T2I) generation models (e.g., Imagen~\cite{saharia2022photorealistic}, DALL·E 2~\cite{ramesh2022hierarchical} and Stable Diffusion~\cite{rombach2022high}) are all based on diffusion process via iterative denoising steps. 
Benefiting from a mountain of training data, these methods have achieved great performance, and also attract tremendous attention from both academia and industry.

\begin{figure}
    \centering
    \scriptsize
    \includegraphics[width=\linewidth]{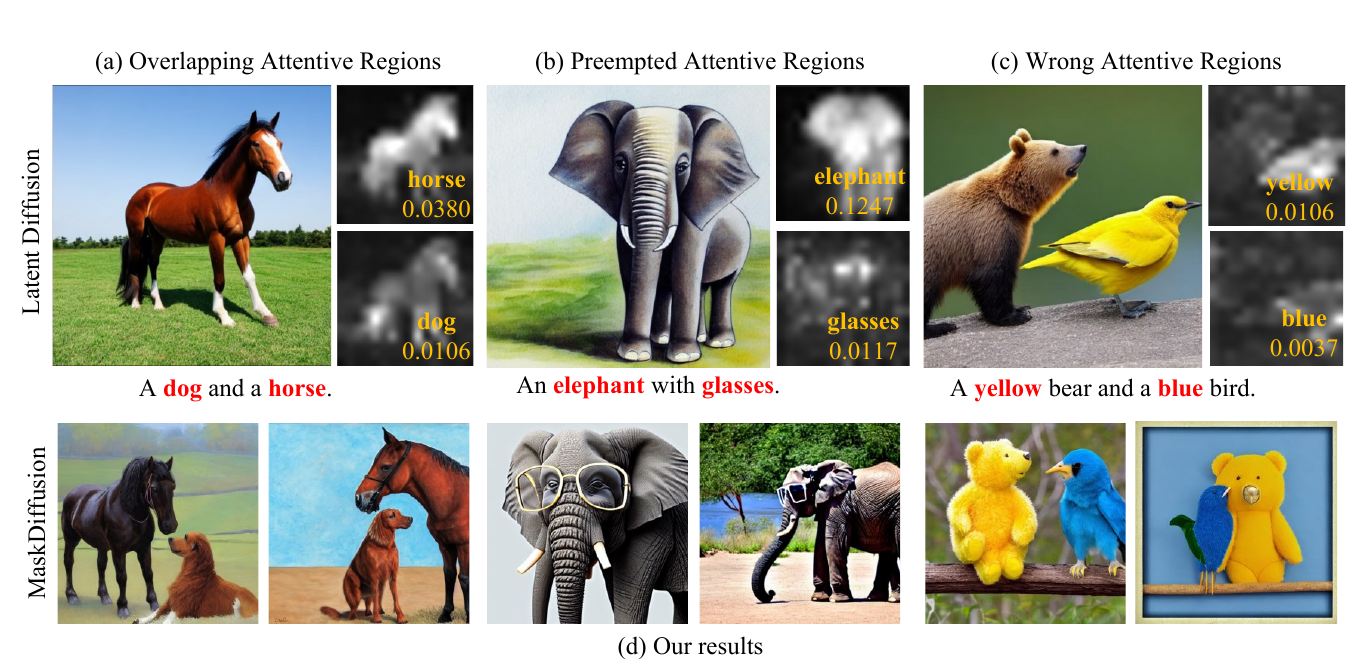}
     \vspace{-10pt}
    \caption{Different types of bad cases in Stable Diffusion and the corresponding cross-attention maps. The numerical score for each cross-attention map is computed by averaging the top 25\% attention values. (a) Both ``horse'' and ``dog''  do not have a semantic relationship but
    suffer from overlapping attentive region problems, which leads to the ``dog'' neglect. 
    (b) There is a large difference between the attention scores of ``elephant'' and ``glasses'', leading to the missing of ``glasses.'' (c) The attentive regions of ``Yellow'' appear at the wrong position, resulting in the generation of a yellow bird. (d) Our results.}
    \label{fig:badcase}
\end{figure}

In the context of T2I diffusion models, despite the great progress, the inconsistency between the text prompt and the generated image is a severe problem~\cite{chefer2023attendandexcite}.
To address this issue, lots of previous works~\cite{saharia2022photorealistic, ramesh2022hierarchical, balaji2023ediffi} scale up diffusion models along with text encoders to achieve better performance but bring in huge extra computational burden.
In this paper, to advocate the sustainable and green AI principle, we follow the principle that \emph{``Entities must not be multiplied beyond necessity.''} 
Our goal is to resolve the drawbacks of existing diffusion models, such as missing objects and mismatched attributes (see \figref{fig:badcase}), without introducing additional computational cost and extra data.

We get inspiration from the prompt-to-prompt work~\cite{hertz2022prompt}, which shows that cross-attention is the key component bridging the text prompt and the generated image. 
This motivates us to delve into the cross-attention mechanism to find a solution to the problem of text-to-image inconsistency as mentioned above. 
Through revisiting the cross-attention mechanism, we find that the main reason preventing T2I diffusion models from generating consistent images with the given prompt is the inadequate cross-modality relation learning.
For example, as shown in the visualized cross-attention maps in~\figref{fig:badcase}, 
the objects or attributes often have attentive regions, but they cannot successfully appear in the resulting images.

To make the resulting image better match the text prompt, we propose \MyMthd{}, a training-free method that can boost the text-to-image consistency.
\MyMthd{} renovates the vanilla cross attention by introducing  a mask to i) prevent the semantic competition (See~\figref{fig:badcase}) in the cross attention maps,  and ii) balance the attention value differences of different tokens.
We conduct comprehensive experiments with Stable Diffusion~\cite{rombach2022high} and compare them with four recent state-of-the-art methods. 
The results reveal the superiority of \MyMthd{} over other methods in generating images highly consistent with the text prompt.

\section{Related Work}

Diffusion models~\cite{rombach2022high, saharia2022photorealistic, ho2020denoising, ramesh2022hierarchical, Peebles2022DiT} 
are predominate in the context of image generation, 
outperforming GANs~\cite{goodfellow2020generative, esser2020taming, kang2023gigagan}, VAEs~\cite{Parmar_2021_CVPR, NIPS2017_7a98af17, NEURIPS2019_5f8e2fa1}, 
and normalized flows~\cite{dflow2021, chen2019residualflows} in fidelity and diversity. 
Since the classifier-free guidance~\cite{ho2022classifierfree} and the conditioning mechanism based on cross attention~\cite{rombach2022high} were proposed, 
conditional generation under diverse types of guidance becomes practical, benefiting to numerous related tasks, 
including super resolution~\cite{saharia2021image}, inpainting~\cite{Lugmayr_2022_CVPR, xie2022smartbrush}, 
layout-to-image generation~\cite{zheng2023layoutdiffusion, cheng2023layoutdiffuse} 
and text-to-image (T2I) generation~\cite{saharia2022photorealistic, ramesh2022hierarchical, balaji2023ediffi}.
Amongst these tasks, T2I diffusion models attract the most significant attention, 
leading to creative and aesthetic applications~\cite{mou2023t2iadapter, zhang2023adding}.

For T2I diffusion models, a core demand is the consistency between the prompts and the generated images. 
However, the existing T2I diffusion models fail to maintain such consistency in some cases~\cite{saharia2022photorealistic}, 
especially when the prompt is extremely long or involves multiple concepts, hindering further application.
To alleviate this issue, Composable Diffusion~\cite{liu2022compositional} makes the first step, 
which takes advantage of the characteristics of classifier-free guidance and 
enables limited operators (i.e., and, not) to achieve better text-to-image alignment.
However, its ability is highly constrained by the limited operators, and it is unsuitable for a fine-grained adjustment.
More recent works can be partitioned into two groups.
They either mine semantic correspondence via cross attention or utilize extra guidance (e.g., layout, silhouette)
with or without finetuning the input latent code. 

In term of extra guidance, most of works~\cite{wu2023harnessing, chen2023trainingfree, ma2023directed} 
utilize layout provided by humans as prior. 
The layout, as a strong prior of position, leads to a strong ability to locate multiple objects of these methods.
Additionally, ~\cite{wu2023harnessing} trains a layout predictor to handle cases while the layout mentioned above is absent.
However, when the number of objects is large, it is burdensome to assign a layout for each one,
and training another network is not an optimal solution as well.

Another category of methods are derived from prompt-to-prompt~\cite{hertz2022prompt}, 
which figures out the strong connection between the generated object and the cross-attention map. 
Attend-and-Excite~\cite{chefer2023attendandexcite} utilizes 
cross attention to optimize the latent code on the fly, 
whose performance is promising but speed is slow, due to the optimization.
Instead, our method is plug-and-play without any latent optimization, 
which only modifies the cross-attention map during the denoising process.
Some concurrent works share a similar idea about attention layer manipulation. 
A brief comparison between our method and theirs is presented below, 
and the detailed technique differences are discussed in \secref{sec:method}.
The Structure Diffusion~\cite{feng2023trainingfree} modifies the value in the \textit{qkv} attention operation and 
needs substantial extra queries to the text encoder, resulting in a surge of computational consumption.
The aim of~\cite{mao2023trainingfree} is to guarantee the size and position of objects, 
while our method focuses on missing objects and attribute mismatch.
Attributing to our laborious designs, 
Our method solves the consistency issue with minimal consumption increase and higher efficiency.

\section{Method}
\label{sec:method}

Our objective is to address the issues presented in current diffusion models, like missing objects and mismatched attributes, without extra training.
Prior to delving into the methodology for enhancing the conformity of generated images with the provided text prompt, it is imperative to revisit the concept of cross-attention within text-to-image diffusion models.

\subsection{Revisiting Cross-Attention}
\label{ssec:revisit}

The cross-attention in Stable Diffusion establishes a connection between the $L$-token encoded prompt $P \in \mathbb{R}^{L\times D}$ and $N$-pixel intermediate noise $\boldsymbol{x}_t \in \mathbb{R}^{N\times D}$ by computing the similarities between the text tokens and the image features as follows:
\begin{equation}
\operatorname{Attention}\left(x_t,P\right) = \operatorname{Softmax}\left(\frac{Q(x_t) K(P)^T}{\sqrt{D}} \right),
\end{equation}
where $Q$ and $K$ are linear transformations and $D$ is the channel dimension. 
For simplicity, we use $C \in \mathbb{R}^{N\times L}$ to denote the resulting attention map. 
Though attention maps can reflect where an object will be located as demonstrated in \cite{chefer2023attendandexcite}, they cannot guarantee the text-to-image consistency.

Given a picked token set, we expect to visualize the cross attention map for each picked token. 
For each picked $token_i$, we extract the cross attention map $C_i \in \mathbb{R}^{N}$  from $C \in \mathbb{R}^{N\times L}$ at each step and average all the cross attention maps to visualize the attentive regions.
After analyzing the bad cases of Stable Diffusion and the corresponding cross attention maps, we find that the attention values in the attention maps are closely related to text-to-image consistency. 
Basically, the common bad cases can be categorized into three types, which are shown in the \figref{fig:badcase} and summarized below:

\begin{itemize}[leftmargin=*]
    \item \textbf{Overlapping Attentive Regions.} 
   When two objects without a semantic relationship overlap in the attention maps and their attention values are close, there will be semantic conflict.
   In this case, the diffusion model will ignore one of the objects. For example, in \figref{fig:badcase}(a), there is no semantic relationship between horse and dog, and hence the diffusion model ignores dog and only generates horse.

    \item \textbf{Preempted Attentive Regions.} In this case, the attentive regions of two objects appear at correct positions but there is a large difference between their attention values.
    For example, in \figref{fig:badcase}(b), the attention score of ``elephant'' is much higher than that of ``glasses.''
    Consequently, ``glasses'' does not appear in the generated image.
    \item \textbf{Wrong Attentive Regions.}
    This means that the attribute appears at a wrong position, which makes the attribute be expressed on other objects.
    As shown in \figref{fig:badcase}(c), the ``yellow'' attribute is reflected on the ``bird'' object. 

\end{itemize}

\subsection{Conditional Mask for Text-to-Image Consistency}
\label{ssec:add}

Regarding the aforementioned analysis, we propose a simple and intuitive solution by adding a mask conditioned on the text prompt embeddings and the cross-attention maps. 
Given the intermediate features $x_t$ and encoded prompt $P$, we reformulate the cross-attention via the following formula:
\begin{equation}
\operatorname{CrossAttn}\left(x_t,P\right) = \operatorname{Softmax}\left(\frac{Q(x_t) K(P)^T}{\sqrt{D}} +  M  \right)V(P)
\end{equation}
where $M \in \mathbb{R}^{N\times L}$ is a mask.
Now, we describe how to generate the mask.

\begin{figure}
    \centering
    \scriptsize
    \includegraphics[width=\linewidth]{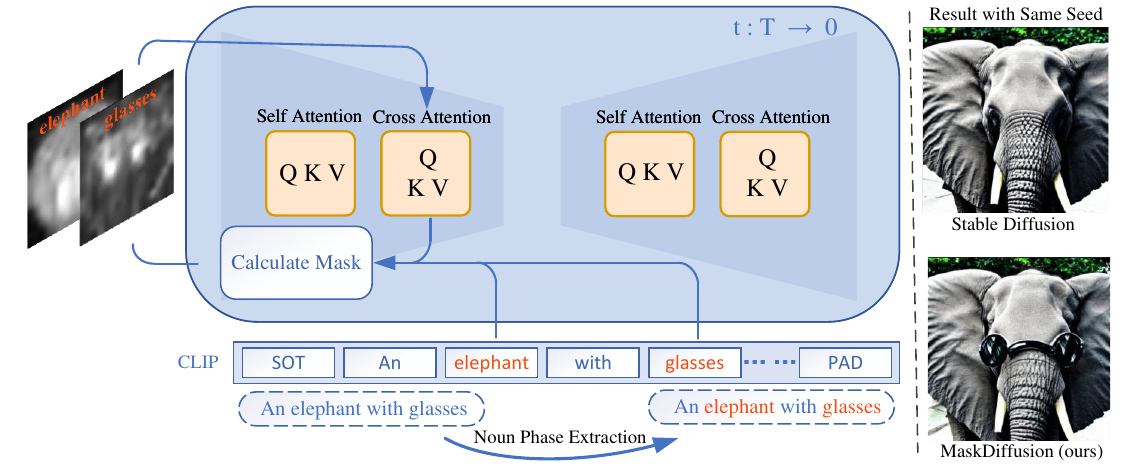}

    \caption{The overall pipeline of \MyMthd{}. The object tokens and their attributes are extracted 
 from prompts before the diffusion process. For each diffusion sample time step, we  calculate a mask of cross attention for the object tokens and their attributes to improving the image-text consistency.}
    \label{fig:overall}

\end{figure}

\begin{algorithm}
	\renewcommand{\algorithmicrequire}{\textbf{Input:}}
	\renewcommand{\algorithmicensure}{\textbf{Output:}}
	\caption{Adaptive Mask Generation}
	\label{alg1}
	\begin{algorithmic}[1]
            \REQUIRE the prompt $P \in \mathbb{R}^{L\times D}$, the picked tokens set $S$, a zeroed mask $M \in \mathbb{R}^{N\times L}$, $w_0 = 5$
		\FOR {$token_i$ in set $S$  }
            \STATE Using the region selection algorithm to find pixels related to the token and putting them into region $R_i$ 
                \FOR {$pixel_j$ in $R_i$  }
                \STATE $M_{ji} \leftarrow  M_{ji} + w_0$
                \ENDFOR 
            \ENDFOR
		\ENSURE  Mask $M \in \mathbb{R}^{L\times N}$ which will be used for improving text-to-image consistency
	\end{algorithmic}  
\end{algorithm}

Given a set of picked tokens $S$, we use the token embeddings to strengthen the semantic information of the image features under the help of the mask.
The complete mask generation process is summarized in Algorithm~\ref{alg1}.
The core is a region selection algorithm, which is descrbed below.

\myPara{Region Selection.} The region selection algorithm finds out the pixels from the attention map $C_i \in \mathbb{R}^{N}$ associated with a certain $token_i$, which is the key of adaptive mask generation. 
We define the region selection process as an optimization problem.
It helps select regions with high response values and meanwhile avoid the overlapping problem as mentioned in \secref{ssec:revisit}. 
In mathematical form, we select regions $R_1,R_2,...,R_S$ for the picked tokens $token_1,token_2,...,token_S$ by maximizing the following objective function $F_\theta$: 
\begin{equation}
\begin{aligned}
& \operatorname{max} \left(\sum_{i=0}^S \operatorname{sum}\left (R_i\right)-\lambda \sum_{i=0}^S \sum_{j=0}^S \operatorname{sum}\left( R_i \cap R_j \right)\right), \\
& \begin{array}{r@{\quad}l@{}l@{\quad}l} 
s.t. &  \forall \, \text{pixel}  \in R_i , C_i[\text{pixel} ] \geqslant \tau_i, \quad i=1,2, \ldots, L
\end{array}
\end{aligned}
\end{equation}
where $S$ is the number of picked tokens and $\operatorname{sum}(\cdot)$ means accumulating the attention value of each pixel in its corresponding region.

Finding the optimal solution to this optimization objective is difficult in that its second term contains the interaction of different regions. 
To resolve this, we simplify the second term of the optimization objective and solve an approximate optimization objective as follows.
We first select the top $k$ largest attention values from the cross attention map associated with each token as the approximate regions $A_1, A_2, ..., A_S$ for the picked tokens. 
After that, we simplify the above optimization objective into the following formula by the approximate regions:
\begin{equation}
\operatorname{max} \left(\sum_{i=0}^S \operatorname{sum}\left (R_i\right)-\lambda \sum_{i=0}^S \sum_{j=0}^S \operatorname{sum}\left( R_i \cap A_j \right)\right).
\end{equation}
The new optimization objective eliminates the correlation between regions $R_1,R_2,...,R_S$.
We do not need to consider other regions when solving for some region $R_i$.
Therefore, it can be solved very quickly.
By default, we set the value of $\lambda$ to 0.5.

In practice, we apply a Gaussian filter over the cross-attention maps as done in~\cite{chefer2023attendandexcite}.
Then,  for the cross-attention map associated with each token after Gaussian smoothing, we select positions whose attention values are larger than 0.5 times of
the largest attention value.
The advantages of these operations are two-fold.
First, it can help erase those regions corresponding to other objects with low attention values.
Second, the highest attention values in attention maps associated with different tokens may have a large difference.
For example, in~\figref{fig:badcase}(b), the attention score of ``elephant'' (0.1247) is much larger than that of ``glasses'' (0.0117).
The above operation can help adaptively select proper attentive regions.

\myPara{Updating Cross-Attention Maps with Momentum.} The optimization objective discussed above solely focuses on adjusting the cross-attention map at the current time step, which may result in unstable masked positions across different time steps. 
We solve this issue by introducing a momentum along the time step dimension.
Specifically, for time step $t$, we update $C^{t}$ by combining the cross-attention maps $C^{t}$ and $C^{t+1}$ as follows:
\begin{equation}
C^{t} \leftarrow  \alpha C^{t+1} + \beta C^{t},
\end{equation}
where $\alpha$ and $\beta$ are the coefficients to adjust the effects of two adjacent time steps.
Here, we empirically set $\alpha=0.03$ and $\beta=0.99$.

\myPara{Overall Pipeline.}
Given a text prompt, our approach involves initially extracting all nouns and adjectives corresponding to the objects present within the visual content, utilizing the SpaCy library~\cite{spacy2}. During the sampling process, we employ the pre-trained U-Net~\cite{ronneberger2015u} network from Stable Diffusion for denoising at each time step, while applying a mask to the cross-attention block. Specifically, the mask $M$ is computed using Algorithm \ref{alg1} of the cross-attention map, focusing on selected tokens such as object nouns and their associated attributes.

To optimize computational efficiency during inference, we solely compute the attention mask pertaining to the extracted nouns and adjectives, disregarding other text tokens. The mask is exclusively applied to feature maps with a resolution of $16\times 16$. A detailed analysis of the impact of feature resolutions can be found in Section \ref{ssec:abl} and Figure \ref{fig:ablation_res}.

\section{Experiment}
\label{sec: exp}
\subsection{Implementation Details}
We implement our method based on version 1.4 of Stable Diffusion~\cite{rombach2022high} pre-trained on the LAION-5B~\cite{schuhmann2022laion}. 
Since our method is training-free, we do not update any parameters of Stable Diffusion. 
In all experiments, the size of the image generated by our model and the comparison model is $512\times 512$. Our test prompts include simple sentences and complex sentences. We sample some prompts from MS-COCO~\cite{lin2014microsoft} and  synthesize some prompts according to the specified format based on GPT~\cite{brown2020language}  to further enhance the diversity of our test data, including the following formats: 1) a [animal/object A] and a [animal/object B], like ``A cat and a dog''. 2) a [color/material A][object/animal A] and a [color/material B][object/animal B], like ``a red bus and a white dog''. 3) More complex sentences: interactions involving more than three objects, like ``A happy dog wagging its tail while fetching a stick in a pond''. More details are described in the supplementary material.

\subsection{Model Efficiency}

We conduct an inference time comparison of five models in this subsection.
%
We generate images using the same 50 prompts and random seeds and take the average inference time for a fair comparison.
The five method names and their corresponding inference time are listed in Tab.~\ref{tab:timescore}.

Composable Diffusion~\cite{liu2022compositional} divides prompt into concepts and composes diffusion models with all concepts. Consequently, the denoising model is employed multiple times within a single sampling time step to calculate different concepts, resulting in a time-consuming process.
Structured Diffusion~\cite{feng2023trainingfree}, on the other hand, splits the sentence into several parts and employs the CLIP text encoder to obtain an embedding.
As the CLIP text encoder is queried multiple times, there is a slight increase in inference time. 
Among the five models, Attend-and-Excite~\cite{chefer2023attendandexcite} exhibits the slowest inference speed, primarily due to the iterative optimization within a single sampling time step.
Our model only introduces a mask conditioned on the prompt embeddings and hence achieves almost the same performance as the original Stable Diffusion, which demonstrates the efficiency of our method.

\begin{table}[htp!]
    \setlength\tabcolsep{8pt}
    \renewcommand{\arraystretch}{1.1}
    \footnotesize
    \centering
    \caption{Quantitative comparison of our \MyMthd{} with Stable Diffusion~\cite{rombach2022high}, ComposeDiffusion~\cite{liu2022compositional}, Structure Diffusion~\cite{feng2023trainingfree}, Attend-and-Excite~\cite{chefer2023attendandexcite}. We give the average inference time, user's supporting rate for simple prompts and complex prompts, and CLIP score similarity to fully measure the performance of models.}
    \begin{tabular}{lcccc}
    \toprule
    \multirow{2}{*}{Model} & \multirow{2}{*}{Inference  Time} & \multirow{2}{*}{\makecell{Simple Prompt  \\Supporting Rate}}  & \multirow{2}{*}{\makecell{Complex Prompt \\Supporting Rate}}  & \multirow{2}{*}{CLIP Score} \\ & & & &\\ \midrule
    Stable Diffusion~\cite{rombach2022high} & 81s &  3.10\%  & 9.16\% & 26.16\\  
    ComposableDiffusion~\cite{liu2022compositional}  &  126s & 0.20\% & 2.29\% & 26.59\\ 
    Structure Diffusion~\cite{feng2023trainingfree}   & 88s & 0.58\% & 2.87\% &25.82 \\ 
    Attend-and-Excite~\cite{chefer2023attendandexcite}  & 215s  & 12.96\% & 11.24\% &26.33\\
 \midrule
    \textbf{\MyMthd{} (ours)}  & \textbf{ 81s} & \textbf{83.16}\% & \textbf{74.77}\% & \textbf{26.82}
 \\ 
     \bottomrule 
    \end{tabular}
    
    \label{tab:timescore}
\end{table}

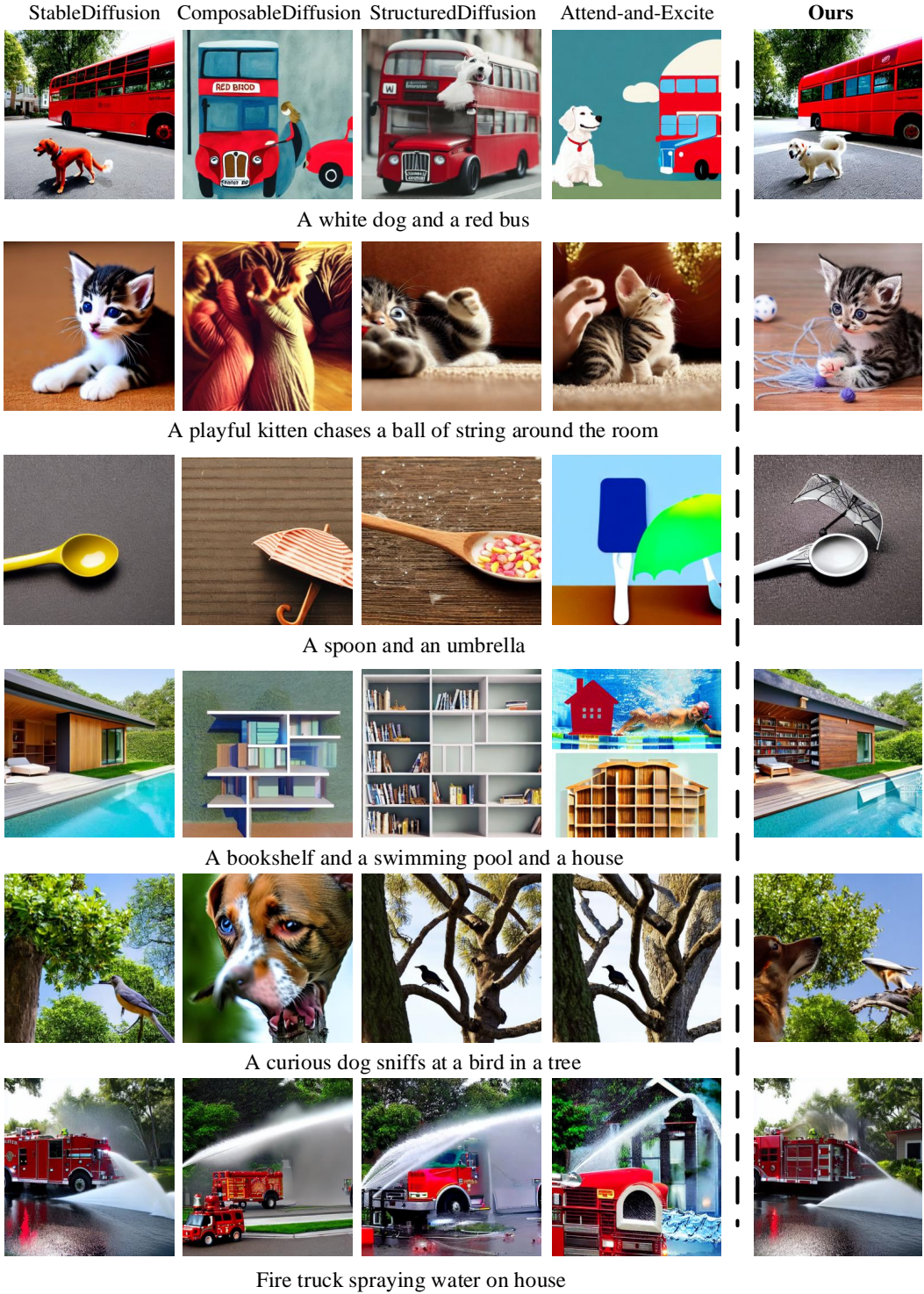
\begin{figure}[!tp]
    \centering
 \footnotesize
    \includegraphics[width=\linewidth]{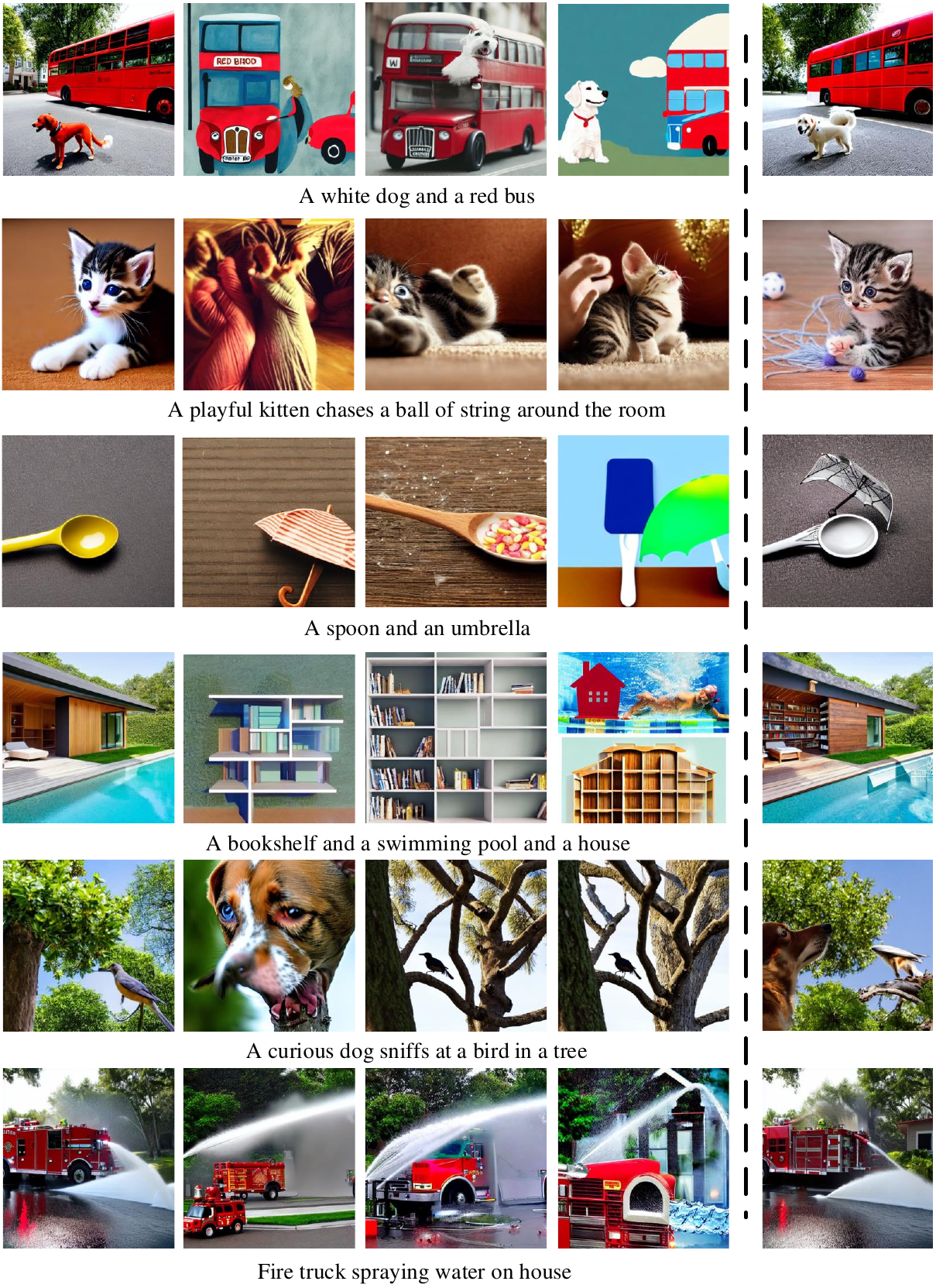}
     \put(-386, 550){StableDiffusion}
     \put(-322, 550){ComposableDiffusion}
     \put(-239, 550){StructuredDiffusion}
     \put(-157, 550){Attend-and-Excite}
     \put(-50, 550){\textbf{Ours}}
     \vspace{-10pt}
    \caption{Visual comparison of \MyMthd{} with Stable Diffusion~\cite{rombach2022high}, Composable Diffusion~\cite{liu2022compositional}, Structured Diffusion~\cite{feng2023trainingfree}, and Attend-and-Excite~\cite{chefer2023attendandexcite}. It can be seen from the generated images that our method achieves the best performance among all the five models.}
    \label{fig:overall}
\end{figure}

\subsection{Visual Comparison}
\begin{figure}
    \centering
    \scriptsize
    \includegraphics[width=\linewidth]{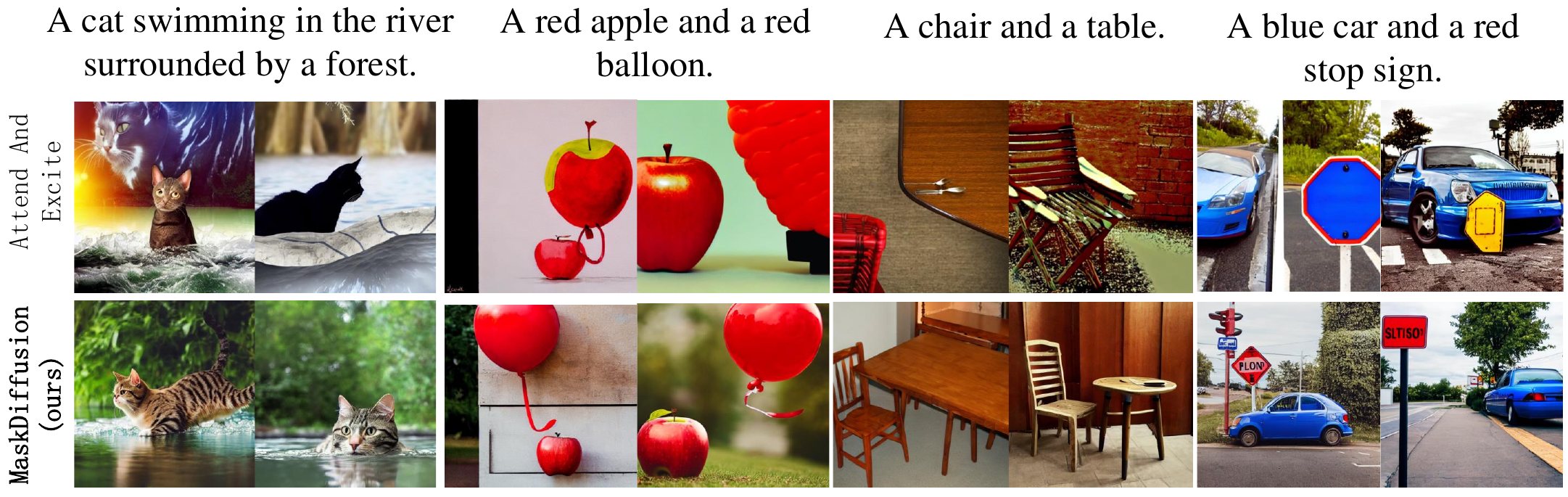}

    \caption{Additional comparisons with the recent Attend-and-Excite~\cite{chefer2023attendandexcite} using the same set of seeds for each prompt. We can see that our method can achieves a better consistency between the text prompt and the generated images.}
    \label{fig:extracompare}
\end{figure}

We compare our model with other four diffusion-based generative models, including Stable Diffusion~\cite{rombach2022high}, Composable Diffusion~\cite{liu2022compositional}, Structured Diffusion~\cite{feng2023trainingfree}, and Attend-and-Excite~\cite{chefer2023attendandexcite}.
We evaluate the matching degree of images generated by different models and prompts using the same seed.
The prompts and generated images are shown in Fig.~\ref{fig:overall}.
Because Attend-and-Excite is the most competitive comparison model among them, we perform an additional comparison with it in \figref{fig:extracompare}
Through a visual comparison of the generated images, we can see that our model produces the most realistic and text-to-image consistent results.
We can also observe that our results are very similar to those produced by the vanilla Stable Diffusion and improve the consistency between the prompts and images.
More examples can be found in our supplementary materials.

\subsection{CLIP Score Comparison}

Here, we use CLIP score similarity to quantitatively evaluate the performance of our model. 
Considering the limitations of CLIP score for semantic understanding~\cite{thrush2022winoground}, we remove sentences that are extremely long or contain complex interactions between multiple objects during evaluation. 
For a fair comparison, we conduct experiments with multiple random seeds and calculate the CLIP score similarity of five models. 
The results are shown in \tabref{tab:timescore}.

\subsection{User Study}
\label{ssec:us}

Here, we conduct a user study to analyze the fidelity of the images generated by our \MyMthd{} and four recent methods.
We randomly sample $30$ prompts and use the same random seed to generate images using the five models. 
The selected prompts include $50\%$ simple prompts, containing two words and attributes, and $50\%$ complex prompts, containing interactions of more than three objects.
The order of the models is shuffled and the participants were not informed which images are generated by some certain model for a fair comparison.

We asked $35$ participants to choose the most similar image to the prompt based on three points: 1) Whether the objects mentioned in the prompt appear. 2) Whether the attributes of the object are correct. 3) Whether the generated image is real and natural. 
We count the users' votes and define it as the user's support rate.
We show the results in \tabref{tab:timescore}. 
As can be seen, our \MyMthd{} can achieve overwhelming user approval no matter how complex the sentences are.

\begin{figure}[htp!]
    \centering
    \scriptsize
    \includegraphics[width=\linewidth]{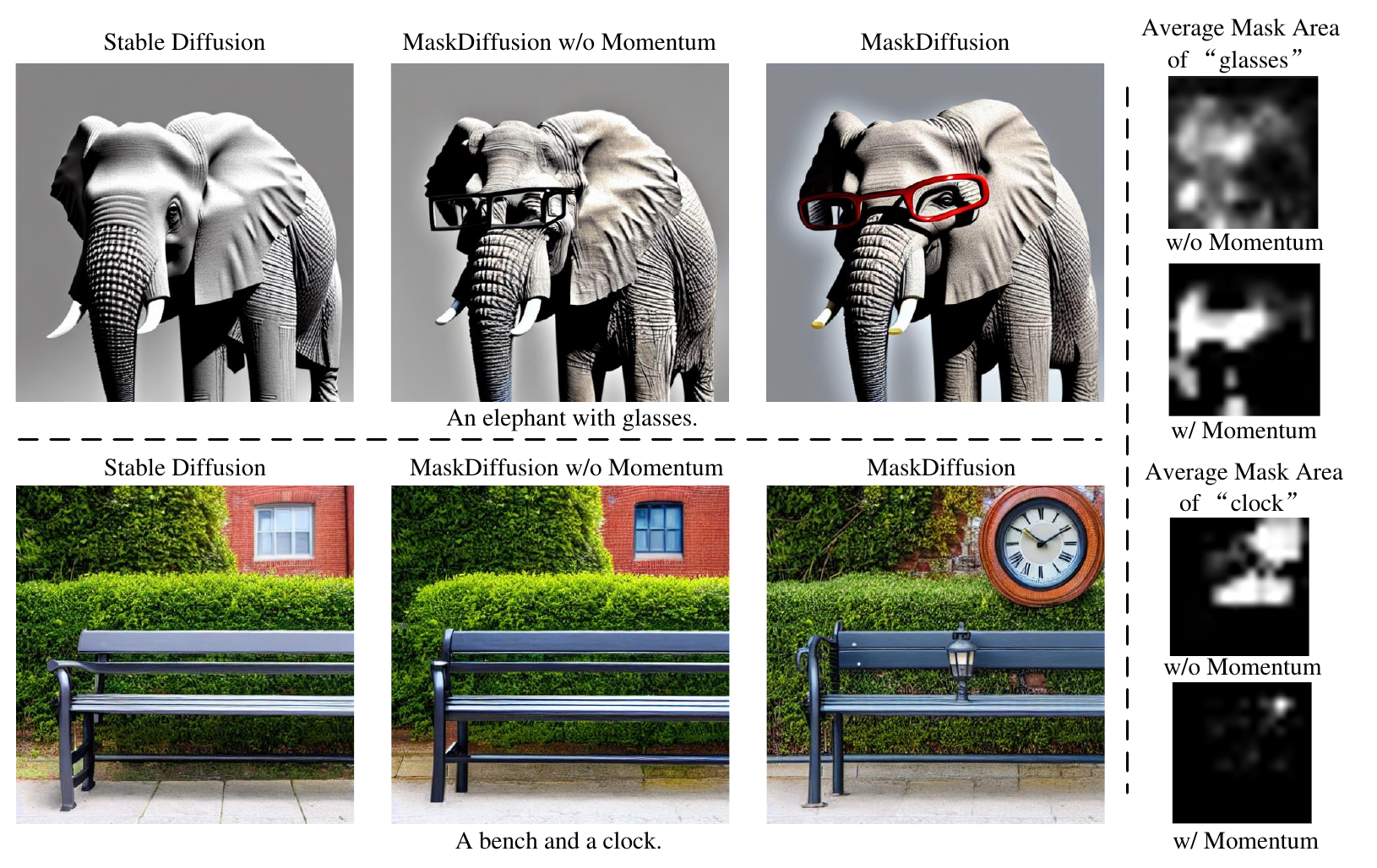}
    \vspace{-15pt}
    \caption{Ablation analysis on the momentum updating strategy. We compare with the original Stable Diffusion and our \MyMthd{} without momentum updating.}
    \label{fig:ema}
\end{figure}

\subsection{Ablation Study}
\label{ssec:abl}

In this subsection, we conduct some ablation analysis to help readers better understand the proposed method.

\myPara{Updating Cross-Attention Maps with Momentum. } We update cross-attention maps with momentum as explained in~\secref{ssec:add}. 
It takes the information of the previous time step into consideration to make the generated mask more stable.
To show the effect of this operation, we design a set of ablation experiments, namely the original Stable Diffusion, \MyMthd{} without momentum updating, and our \MyMthd{}, and show the results in~\figref{fig:ema}.
We also record the mask at each time step and compute the averaged mask for each token in the right part of~\figref{fig:ema}.
From the figure, we can see that without the momentum updating strategy, some of the objects may be missing in the resulting images.
On the contrary, our \MyMthd{} can well address this issue.

\begin{figure}
    \centering
    \scriptsize
    \includegraphics[width=\linewidth]{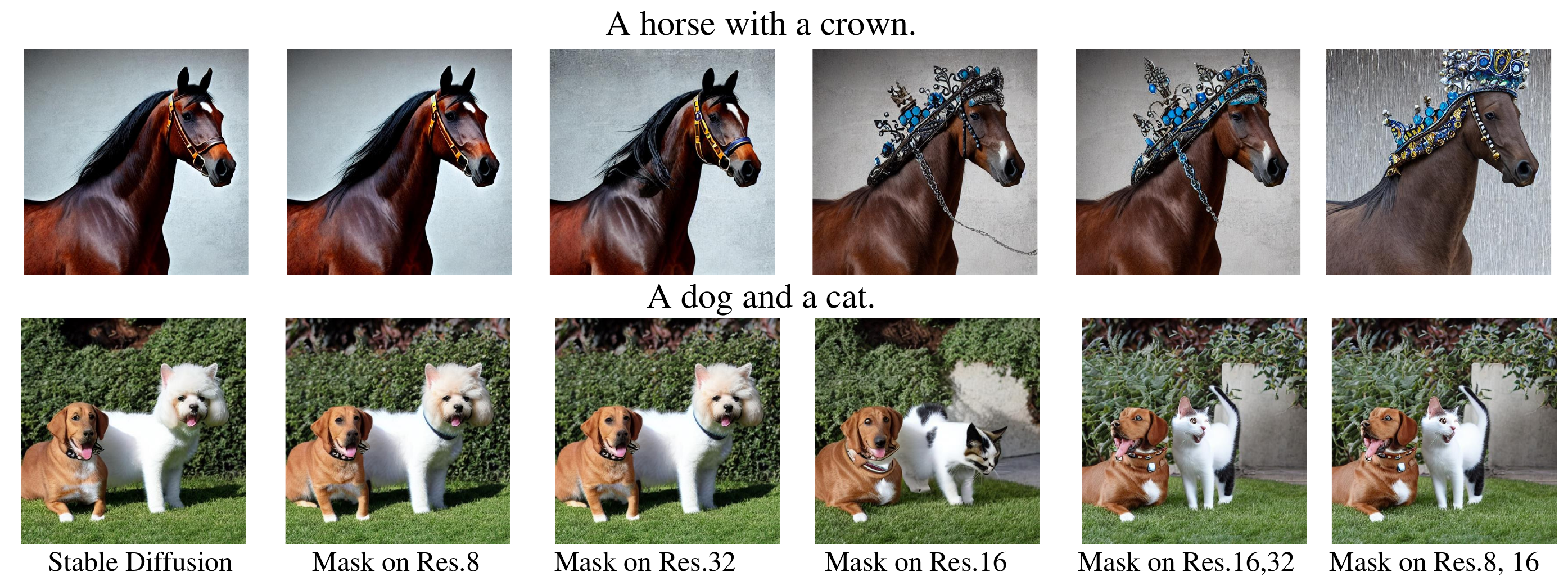}
    \caption{Ablation study of adding mask on cross attention of different resolution features.  ``res.'' is an abbreviation for resolution. For example, ``res.8'' stands for $8\times8$ resolution U-Net features. 
    It can be clearly seen that generating a mask based on the $16\times 16$ features plays a decisive role in the success of our \MyMthd{}.}
    \label{fig:ablation_res}
\end{figure}

\myPara{Masks Generated at Different Resolutions. } The diffusion model adopts a hierarchical structure, i.e., the UNet. In Stable Diffusion, the cross-attention is computed based on features at each feature level. 
In order to clarify which resolution is more appropriate to add masks, we conduct ablation experiments by adding masks to cross-attentions of different resolutions.
As shown in~\figref{fig:ablation_res}, we can observe that the $16\times 16$ resolution plays an important role in the success of our method. 
A reasonable explanation is that during CLIP training, the text and image embeddings are aligned. 
The resolution of the output image embeddings in CLIP is $14\times14$ and the $16\times16$ resolution features in U-Net is close to it.

\section{Conclusions}

In this paper, we have conducted extensive research on the image-text mismatch problem of the text-to-image diffusion model, revealing the correlation between the inappropriate attention value of the cross-attention map and the image-text mismatch problem. To amend the inappropriate attention value,  we introduced an adaptive mask to control the attention value of the cross-attention map, conditioned on the cross-attention map and prompt semantics.
Based on this, we propose a simple and training-free method called \MyMthd{} to generate high-quality images with high text-to-image consistency. 
Under extensive experiments, our \MyMthd{} achieves 
better text-to-image consistency with low cost compared to the original diffusion models.
We believe that our work provides a valuable contribution to the field of text-to-image synthesis. we hope that it can be used to improve the performance of future text-to-image synthesis models.

\myPara{Limitations.} 
The main limitation of our method arises from the CLIP text encoder~\cite{radford2021learning}. Our method can adaptively add masks based on the cross-attention map and CLIP semantics, while the descriptive capacity of the CLIP text encoder for complex sentences may be limited, leading to ambiguous or incorrect semantics. 
For instance, when evaluating the prompt ``A green book in a pink basket.'', we found that CLIP misunderstands the sentence, as evidenced by the higher cosine similarity between ``basket'' and ``green'' compared to ``basket'' and ``pink.'' The ambiguous semantics can confuse the diffusion generation process, which we plan to address in our future work.

{\small
\bibliographystyle{ieee_fullname}
\bibliography{egbib}
}

\end{document}